\documentclass[sigconf]{acmart}
\usepackage{booktabs} % For formal tables
\usepackage{multirow}
\usepackage[lined,algoruled,linesnumbered,noend]{algorithm2e}
\usepackage[noend]{algpseudocode}
\algdef{SE}[DOWHILE]{Do}{doWhile}{\algorithmicdo}[1]{\algorithmicwhile\ 
	#1}
\usepackage{footnote}
\usepackage{tablefootnote}
\makesavenoteenv{tabular}
\newcommand{\statementDef}{(\entity{s}, \entity{p}, \entity{o})}

\newcommand{\entity}[1]{\textsf{#1}}

	\newcommand{\kgDef}{G\!\!=\!\!(V,E,T)}
\newcommand{\TBoxDef}{T\!\!=\!\!(C_t, P_t,L_t,T_t)}
\newcommand{\kgraph}{KG}
\newcommand{\kgraphs}{KGs}
\newcommand{\system}{FEA}
\newcommand{\factDef}{(\entity{s}, \entity{p}, \entity{o})}

	\newcommand{\kgSchemaDef}{G_S\!\!=\!\!(V_s,E_s,T_s)}

\newcommand{\setPaths}{{\mathcal{P}\factDef=\{\pi_1,\pi_2,\ldots,\pi_k\}}}
\newcommand{\fact}[3]{(\entity{#1}, \entity{#2}, \entity{#3})}
	\newcommand{\allEvPatterns}{\mathcal{P}}
	\newcommand{\domains}{\mathcal{D}oms} 
\newcommand{\ranges}{\mathcal{R}angs} 
\newcommand{\targetP}{{ \entity{p}}}

\newcommand{\rightPathElement}[3]{ {\small \entity{#1}\xrightarrow{\text{#2}} \entity{#3}}}
\newcommand{\leftPathElement}[3]{ {\small \entity{#1}\xleftarrow{\text{#2}} \entity{#3}}}
\newcommand{\emb}[1]{{ \entity{Emb}(#1)}}
\newcommand{\embF}[1]{{ \entity{EmbF}(#1)}}
\usepackage{pgfplots}
\usepackage{pgfplotstable}
\usetikzlibrary{plotmarks}
\usepackage{pgfgantt}

\usepackage{multirow}

\usepackage{balance}
\def\addlegendimage{\csname pgfplots@addlegendimage\endcsname}
\pgfplotsset{compat=1.13}
\definecolor{Gray}{gray}{0.95}
\definecolor{LGray}{gray}{0.75}
\definecolor{bblue}{HTML}{4F81BD}
\definecolor{rred}{HTML}{C0504D}
\definecolor{ggreen}{HTML}{9BBB59}
\definecolor{ppurple}{HTML}{9F4C7C}
\usetikzlibrary{shapes.geometric,arrows.meta} %<added arrows.meta
\tikzset{vertex style/.style={
		draw=#1,
		thick,
		fill=#1!70,
		text=white,
		ellipse,
		minimum width=2cm,
		minimum height=0.75cm,
		font=\small,
		outer sep=0.3pt,
	},
	text style/.style={
		sloped,
		text=black,
		font=\footnotesize,
		above
	}
}
\tikzset{
	vector/.pic={
		\draw (0,0) rectangle (0.5,0.25) rectangle (1,0) rectangle (1.5,0.25) rectangle (2,0); %rectangle (2.5,0.5);
	}
}

\acmConference[Draft'2020]{Draft}{2020}{} 
\acmYear{2020}
\copyrightyear{2020}
\acmPrice{00.00}

\newcommand\HUGE{\@setfontsize\Huge{90}{120}} 

\begin{document}
\title{Fact-checking via Path Embedding and Aggregation}
%\titlenote{Produces the permission block, and
 % copyright information}
%\subtitle{Extended Abstract}
%\subtitlenote{The full version of the author's guide is available as
  %\texttt{acmart.pdf} document}

\author{Giuseppe Pirr\`o}
%\authornote{Dr.~Trovato insisted his name be first.}
\affiliation{%
  \institution{Sapienza University of Rome, Italy}
 % \streetaddress{P.O. Box 1212}
  %city{Dublin} 
}
\email{pirro@di.uniroma1.it}

%\authornote{The secretary disavows any knowledge of  author's actions.}
%\affiliation{%
%  \institution{Institute for Clarity in Documentation}
%  \streetaddress{P.O. Box 1212}
%\email{webmaster@marysville-ohio.com}

\begin{abstract}
Knowledge graphs (\kgraphs) are a useful source of background knowledge to (dis)prove facts of the form \fact{s}{p}{o}. Finding paths between \entity{s} and \entity{o} is the cornerstone of several fact-checking approaches. While paths are useful to (visually) explain why a given fact is true or false, it is not completely clear how to identify paths that are most relevant to a fact, encode them and weight their importance. 
The goal of this paper is to present the \textbf{F}act checking via path \textbf{E}mbedding and \textbf{A}ggregation (\system) system.  \system\ starts by carefully collecting the paths between \entity{s} and \entity{o} that are most semantically related to the domain of \entity{p}. However, instead of directly working with this subset of all paths, it learns vectorized path representations, aggregates them according to different strategies, and use them to finally (dis)prove a fact. We conducted a large set of experiments on a variety of \kgraphs\ and found that our hybrid solution brings some benefits in terms of performance.
\end{abstract}

%\begin{CCSXML}
%<ccs2012>
%  <concept_id>10010520.10010553.10010562</concept_id>
%  <concept_desc>Computer systems organization~Embedded 
%systems</concept_desc>
%  <concept_significance>500</concept_significance>
%  <concept_id>10010520.10010575.10010755</concept_id>
%  <concept_desc>Computer systems 
%organization~Redundancy</concept_desc>
%  <concept_significance>300</concept_significance>
%  <concept_id>10010520.10010553.10010554</concept_id>
%  <concept_desc>Computer systems 
%organization~Robotics</concept_desc>
%  <concept_significance>100</concept_significance>
%  <concept_id>10003033.10003083.10003095</concept_id>
%  <concept_desc>Networks~Network reliability</concept_desc>
%  <concept_significance>100</concept_significance>
%</ccs2012>  
%\end{CCSXML}
%
%\ccsdesc[500]{Computer systems organization~Embedded systems}

%10 pages plus references


\maketitle

\section{Introduction}
\label{sec:intro}
%%%%
We live in a digital era, where both false and true rumors spread at an 
unprecedented speed. In this open context, having a way to assess the 
reliability of individual facts is of utmost importance. 
How could fact-checkers or even simple citizens quickly verify the reliability of statements like  (\entity{Dune}, \entity{directed}, \entity{D. Lynch})? 

One way would be to employ time-consuming techniques requiring to both manually 
collect and check (digital) evidence; for instance, one could look up sources like 
encyclopedias, newspapers and  even gain further evidence by asking 
friends. Another way is to devise automatic fact-checking systems\footnote{https://fullfact.org/blog/2016/aug/automated-factchecking/}~\cite{thorne2018automated,vlachos2014fact}. Existing approaches, can roughly been categorized in three main categories. 
\textit{First}, \textit{text-based approaches} based on a variety of learning models; these can use probability and logics (e.g., \cite{ahmadi2019explainable}), deep-learning (e.g., \cite{karadzhov2017fully}), and also include multi-modal (e.g., text and video) information (e.g., \cite{ar2019mvae}). While these approaches can rely on large amounts of text and/or mutimedia sources like audio and video, there are difficulties in automatically understanding such pieces of information to (dis)prove a fact. \textit{This makes it difficult to give precise semantics to the fact being checked and contextualize it}. On one hand, giving {semantics} boils down to understanding the fact itself rather than relying on statistical indicators like the popularity of a tweet about the fact. For instance, to (dis)prove the fact \fact{Dune}{director}{D. Lynch}, it is crucial to understand that the predicate \entity{director} relates a \entity{Film} and a \entity{Director} and that \entity{Director} is a subclass of \entity{Person}. On the other hand,  contextualizing facts and gaining insights from (chains of) related facts can represent a valuable source of knowledge~\cite{voskarides2018weakly}. As an example, the fact \fact{Jaguar}{owner}{Tata Motors} provides more insights when understanding that it is about the car brand instead of the animal; the additional fact \fact{Tata Motors}{type}{Company} can help in shedding light on this aspect.

%%%%
%%%%
\textit{Second}, \textit{approaches that leverage structured 
	knowledge} (e.g., knowledge graphs) instead of unstructured text (e.g., \cite{shi2016fact,fionda2018fact,pan2018content,shiralkar2017finding}). In this case, structured background knowledge allows for more precise forms of reasoning for fact-checking. For instance, it has been shown that the paths between the \entity{subject} and \entity{object} of a targeted fact, that include other entities and predicates, form a valuable body of semantic evidence (see e.g., \cite{shi2016discriminative, fionda2018fact}). These approaches offer advantages in terms of semantic interpretation and contextualization of a statement. For instance, the statement \fact{Dune}{director}{D. Lynch} can be given both a semantic characterization and put into context by looking at the excerpt of \kgraph\ in Fig. \ref{fig:example-graph} (a) taken from DBpedia. Here, we see that the domain of the fact is that of movies and that there are frequently occurring semantic relations between D. Lynch and  actors (e.g., K. Mclaughlin) that also acted in Dune. Moreover, the usage of paths or entire portions of a \kgraph\ of interest for the target statement can provide (visual) evidence about why the fact is true or false. By looking at the evidence depicted in Fig. \ref{fig:example-graph} (a), it becomes plausible to consider the statement \fact{Dune}{director}{D. Lynch} as true. \textit{Nevertheless, \kgraph-based approaches lack mechanisms to automatically differentiate the importance of the collected paths}.
%
%
%%%%
\textit{Third}, a more recent strand of research has considered the usage of 
\textit{entity and predicate embeddings} {for fact-checking (e.g., \cite{shi2017proje, bordes2013translating}). The idea of these approaches is to treat fact-checking as  a link prediction problem. \textit{While these approaches have the advantage of working with vectorized representations of entities and predicates to automatically identify and extract salient features, they are sub-optimal as they do not directly tackle the problem of vectorizing entire facts, paths, and their aggregation}. 

\subsection{Overview} 
The goal of this paper is to present the \textbf{F}act checking via path \textbf{E}mbedding and \textbf{A}ggregation (\textbf{FEA}) system. FEA carefully collects paths from a KG between the subject and object of a fact to be checked that are most semantically relevant to it. However, instead of directly working with this subset of all paths, it learns vectorized path representations, aggregates them according to different strategies, and use them to finally (dis)prove a fact. To the best of our knowledge, this is the first work combining triple and path embedding and aggregation for fact checking.

To outline the rationale of \system, we consider again the statement \fact{Dune}{director}{D. Lynch} and the excerpt of the DBpedia \kgraph\ along with its schema shown in Fig.~\ref{fig:example-graph}. As previously mentioned, the (visual) semantic evidence in  Fig.~\ref{fig:example-graph} (a) helps in understanding that the fact is true; we can see, for instance, that \entity{D. Lynch} directed other movies where the same actors as \entity{Dune} were acting. Besides, we also note that some patterns like $\Big[\rightPathElement{*}{starring}{*}$$\leftPathElement{}{starring}{*}$$\rightPathElement{}{director}{*}\Big]$ or $\Big[\rightPathElement{*}{cinematography}{*}$$\leftPathElement{}{cinematography}{*}$$\rightPathElement{}{director}{*}\Big]$ emerge.
While this analysis is easy from a visual perspective, automatizing it sets two main challenges. The first is about how to extract such patterns and the corresponding paths to find semantic evidence. The second is about how to inject it into an automatic fact-checking mechanism. \system\ tackles these challenges in three steps:
\vspace{-.1cm}
\begin{enumerate}
	\item \textit{Evidence Collection from the Schema}: given the predicate \entity{p} in a target fact $\factDef$, \system\ constructs fact templates from the \kgraph\ schema. Fig.~\ref{fig:path-extraction} (a) shows two of the available templates for the predicate \entity{director} obtained from the schema in Fig.~\ref{fig:example-graph} (b). Then, it finds schema-level patterns from the subject (\entity{Film}) to the object (\entity{Person}) of the fact template that only include the top-$k$ most related to the input predicate \entity{p}. As an example, for \entity{director}, patterns including \entity{starring}, \entity{cinematography}, and \entity{editing} will be preferred to those including less related predicates as \entity{birthYear} or \entity{college}.
	\item \textit{Evidence Collection from the Data}: \system\ finds data-level paths complying with schema-level patterns found in (1). When using the pair (\entity{Dune}, \entity{D. Lynch}) to replace the endpoints of the patterns in Fig. \ref{fig:path-extraction} (b), \system\ finds the data-level paths {\tiny $\Big[\rightPathElement{Dune}{starring}{J. Nance}$$\leftPathElement{}{starring}{Eraserhead}$$\rightPathElement{}{director}{D. Lynch}\Big]$} and \\{\tiny $\Big[\rightPathElement{Dune}{cinematography}{F. Francis}$$\leftPathElement{}{cinematography}{The E. Man}$}\\{$\rightPathElement{}{director}{D. Lynch}\Big]$}, among the others. 
	\item \textit{Learning}: %\system\ to be able to reason over paths using deep-learning techniques, paths are first given a vectorized representation.
	 \system\ represents each fact \entity{t}=\fact{s$_i$}{p$_j$}{o$_k$} in a path as  a vector $t_E$=\embF{t}, where \embF{$\cdot$} is a fact embedding function.
	Then, a path is represented as a sequence of such vectors. \system\ learns an aggregate representation of all paths $\mathcal{P}_V$ according to different strategies (detailed in Section~\ref{sec:aggregators}), ranging from simple ones that average the contribution of each path (\textit{AvgPooling}) to more sophisticated ones able to also capture dependencies between facts in a path (\textit{LSTMMaxPool}). A final verdict about a targeted fact is provided by giving as input to a classifier the aggregate representation of all paths $\mathcal{P}_V$ (Section~\ref{sec:fact-checker}).
\end{enumerate}
\subsection{Contributions}
\label{sec:contributions-outline}
In this paper, we contribute a fact-checking approach that 
given a fact of the form \statementDef\ assigns a truth score. We make the following main contributions:
\begin{enumerate}
	\item A \textit	{schema-driven algorithm} for extracting paths 
	from \kgraphs \ that can provide evidence to (dis)prove a fact;
	\item An approach to \textit{vectorize paths} based on the embeddings of facts in a path.
	\item Different \textit{path aggregation strategies} to assemble semantic evidence from paths useful for fact-checking.
	\item An extensive evaluation on a variety of datasets and a comparison with related work.
	 \end{enumerate}
 %%%&

Our approach delivers better performance (in terms of number of facts correctly evaluated) than the state-of-the-art (e.g.,~\cite{fionda2018fact,shiralkar2017finding}), while at the same time giving more flexibility in terms of strategies to  collect semantic evidence in the form of paths, distillate and feed such evidence to a learning model.
\begin{figure}[!h]
	\centering
	\includegraphics[width=\columnwidth]{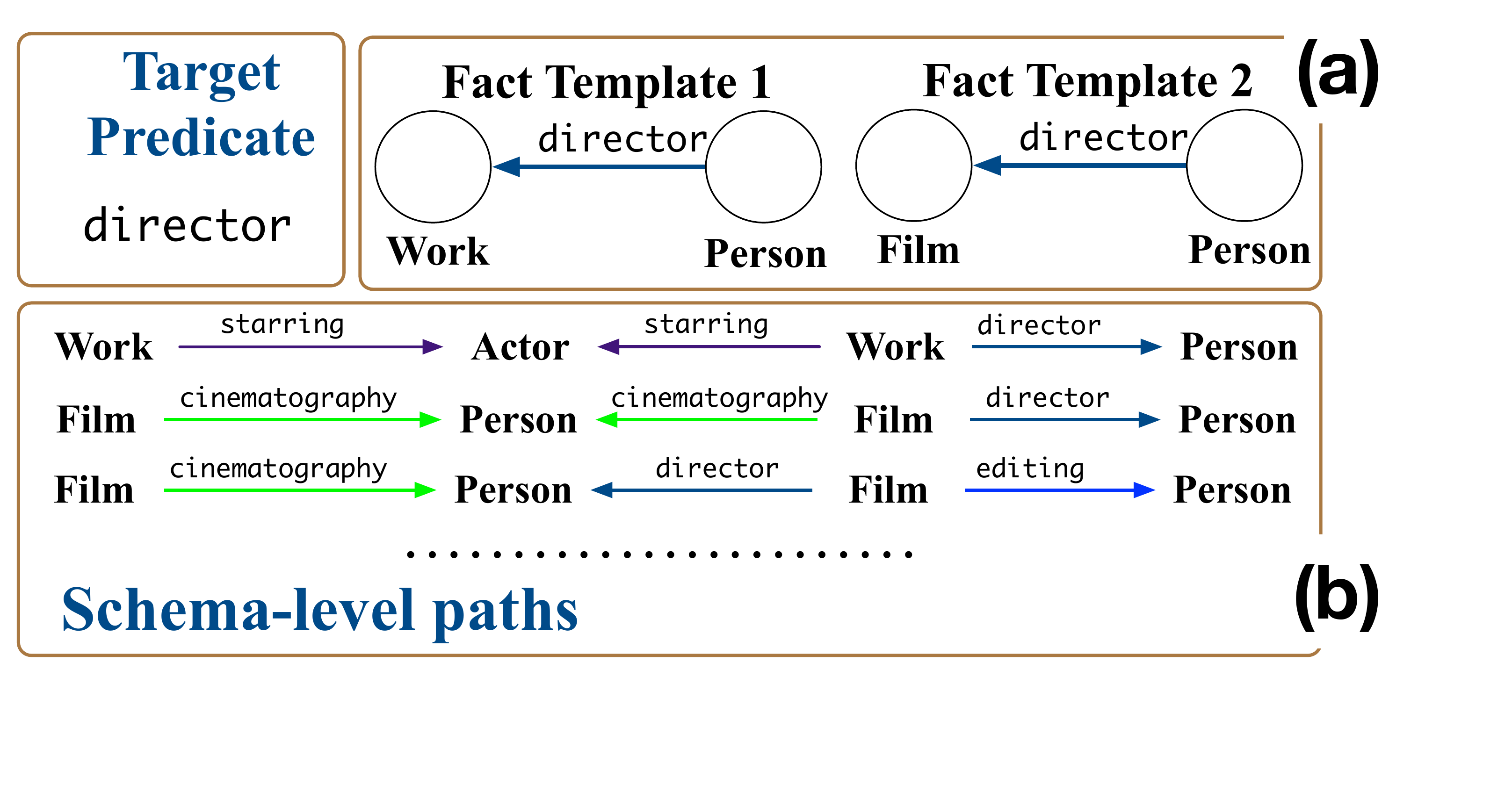}
	\vspace{-1.3cm}
	\caption{Fact templates and schema-level patterns.}
	\label{fig:path-extraction}
\end{figure}
%%%%%%%%
\section{Preliminaries}
\label{sec:preliminaries}
A Knowledge Graph (\kgraph) contains facts (aka statements) that can be divided into an ABox and a 
TBox. We see the ABox as a node and edge-labeled directed multi-graph $\kgDef$ 
where $V$ is a set of uniquely identified vertices representing entities (e.g., \entity{D. 
Lynch}), $E$ a set of 
predicates or properties (e.g., \entity{director}) and $T$ a set of facts 
of the form \statementDef, where
\entity{s}, \entity{o} $\in V$ and \entity{p} $\in E$.  
%
%%%%%%%%%%
%%%%%%%%%%
{The TBox is another multi-graph defined as $\TBoxDef$, where $C_t$ is the set of all class names, $P_t$ is the set of all property names, $L_t$ is a set of properties defined in some ontological language, and $T_f$ is a set of triples of the form $(u,p,v)$ where $u, v \in C_t \cup P_t$ and $p \in L_t$}. Fig. \ref{fig:example-graph} (a) shows an excerpt of the DBpedia TBox used to structure knowledge of the ABox in Fig. \ref{fig:example-graph} (a). Here, we can see that \entity{director} has as domain \entity{Film} or that \entity{Artist} is a subclass of \entity{Person}. In this paper, we consider $L_t$ to be the subset of the RDFS ontological language defined as follows: $L_t$=\{\entity{rdfs:subClassOf}, \entity{rdfs:subPropertyOf}, \entity{rdfs:domain}, 
\entity{rdfs:range}\}. We consider RDFS as it is widely available and allows new facts about the TBox to be efficiently 
derived by applying (a subset of) the RDFS inference rules %franconi2013logic
\cite{munoz2009simple}. In what follows, we use the notation $domain(p)$ (resp., $range(p)$) to indicate the domains (resp., ranges) of a property. 
After applying the RDFS inference on the TBox $T_f$, we consider an alternative graphic representation, which facilitates the extraction of schema-level patterns. We call this representation TBox Graph (Fig. \ref{fig:example-graph} (b)).
\begin{definition}\normalfont  {\textbf{(TBox Graph)}}. Given a 
	TBox $\TBoxDef$, its TBox graph is defined as $\kgSchemaDef$, where each $v_i\in V_s$ is a class name belonging to $C_t$, $p_i\in P_t\cup \{\entity{rdfs:subClassOf}\}$, and  $(v_s, p_i, v_t)\in T_s$ is a triple such that $domain(\entity{p}_i)$=$v_s$ and $range(\entity{p}_i)$=$v_t$.
	\label{def:schema-graph}
\end{definition}

\noindent
%
%,
\begin{figure*}[!h]
	\centering
	\includegraphics[width=.75\textwidth]{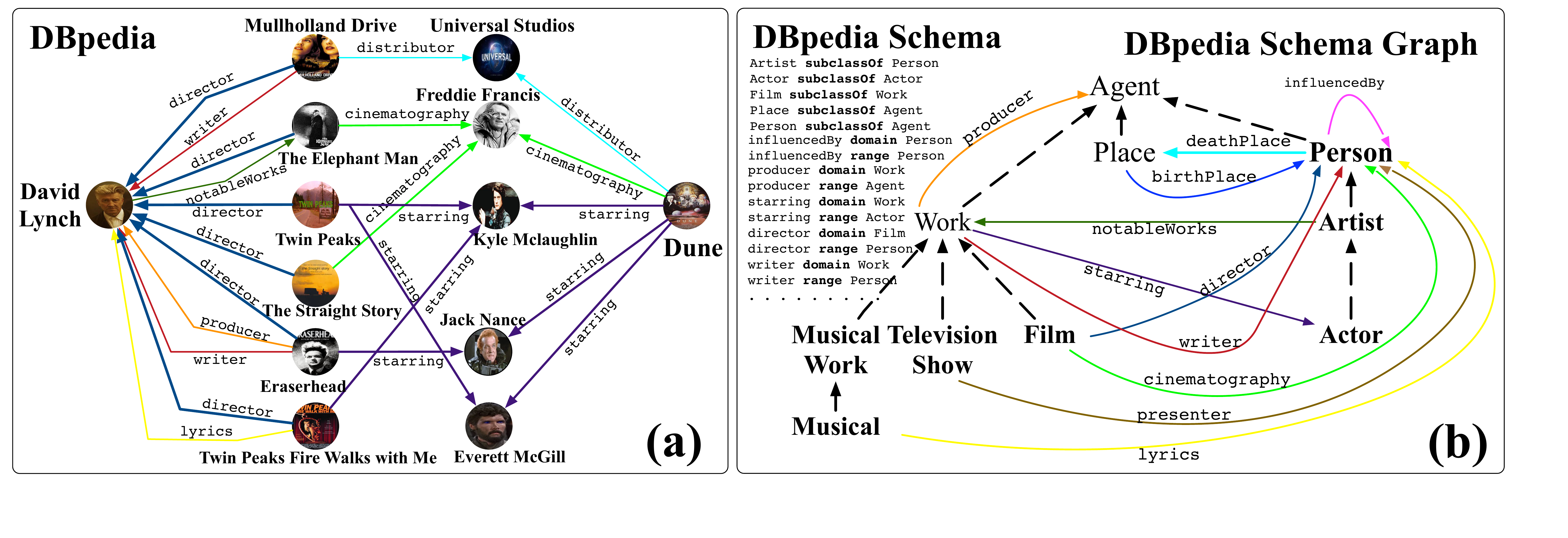}
	\vspace{-1cm}
	\caption{An excerpt of the DBpedia (a), its schema and its schema graph (b).}
	\label{fig:example-graph}
	%
	%\vspace{-.5cm}
\end{figure*}
%, 

\noindent
\textbf{Schema availability and completeness.} Our approach assumes the \kgraph\ to be endowed with a minimal schema definition. This assumption is realistic in practice; popular \kgraphs\ like DBpedia, Yago, and Wikidata feature even richer schema definitions. Another aspect concerns the completeness of the schema as it can be the case that some classes 
and properties are under-specified (e.g., missing domain and range). Our experiments 
on different datasets covering a broad number of domains, show that when the available schema specifications miss domains and ranges, considering the general concept Thing in lieu of them is enough. In general, one can manually refine/complete the portion of the schema that touches a particular domain of interest or use approaches that surface/refine a KG schema from the data or rely on other schema definitions like schema.org or WordNet to complete under-specified aspects in the schema of the \kgraph\ of interest.
\section{The \system\ Framework}
\label{sec:approach-description}
%
%%%
\system\ is an end-to-end framework combining knowledge graph exploration and deep learning techniques for fact-checking. The cornerstone of \system\ to (dis)prove a fact is the ability to both contextualize the fact, collect semantic evidence in the form of paths and use it into a deep-learning model.
The problem that we solve can be formulated as follows: given a fact $\factDef$, and a set of paths $\setPaths$ connecting \entity{s} and \entity{o} and \textit{related to the domain} expressed by \entity{p}, the overall goal is to estimate the truthfulness of the fact by:
\vspace{-.1cm}
\begin{equation}
\Phi_{\factDef}=m_\Theta(\factDef,\mathcal{P}(\factDef))
\end{equation}

\noindent
where $m$ is the model having parameters $\Theta$ and $\Phi \in [0,1]$ is the 
truthfulness score. The \system\ framework consists of four main modules: \textit{path extractor}, \textit{path embedder}, \textit{path aggregator}, and \textit{fact checker}; Fig.~\ref{fig:learning-model} provides an overview of the framework. The end-to-end learning objective of \system\ is guided by the input fact to be checked \factDef\ and a set of (domain-specific) paths extracted from a knowledge graph. The output of the model $\Phi_{\factDef}$ represents the truthfulness of the fact. Note that the set of input paths can provide evidence useful to understand why a given fact is true or false, for instance by displaying paths, similarly to what it has been done in Fig. \ref{fig:example-graph} (a). We now describe each module of the framework.
%%%%%
 \begin{figure}[!h]
	\centering
	\includegraphics[width=\columnwidth]{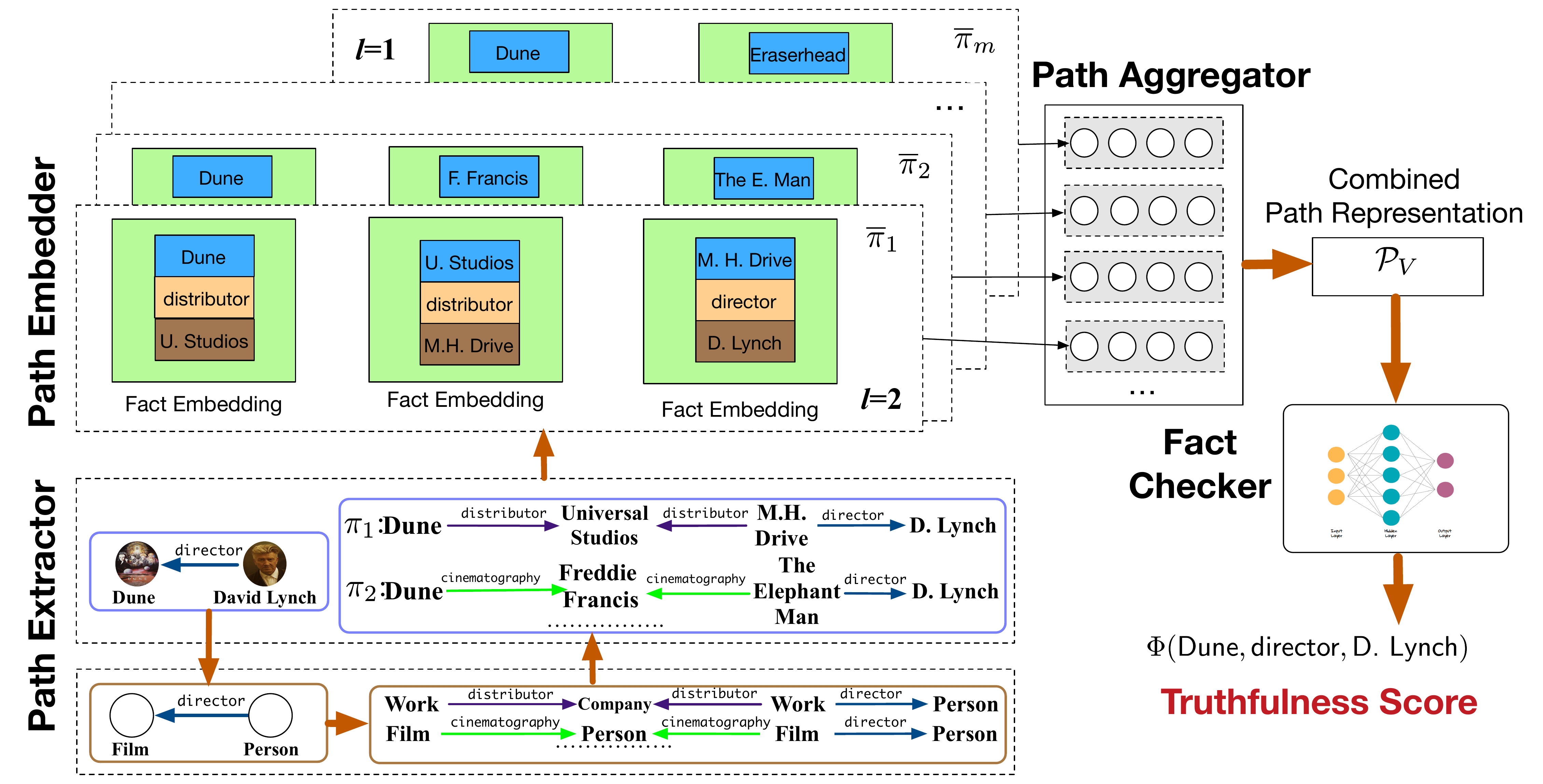}
	\vspace{-.3cm}
	\caption{{Overview of the \system\ framework. Paths are grouped and processed for each length $l$.}}
	\label{fig:learning-model}
\end{figure}
\subsection{Path Extractor}
\label{sec:extract-paths}
This module is responsible for the exploration of the \kgraph\ to gather information in the form of paths, which will be used by the other modules. The Path Extractor does not blindly explore the whole path search space, which can be huge, but focuses on finding paths that are relevant to the input fact via a two step process. First, it explores the  \kgraph's schema conditioned on an input predicate to learn schema-level patterns. Second, it explores the \kgraph's data conditioned on the schema-level patterns and input fact to generate the data-level paths.
%%%
\subsubsection{Schema-level patterns.} 
\label{sec:schema-level-paths}
The Path Extractor leverages the TBox graph to find schema-level patterns for a specific input predicate \entity{p}. It assembles paths up to a length $l$ between the domain(s) and range(s) of \entity{p} treating the input graph as undirected. To reduce the search space, this module only extracts the schema-level patterns most relevant to \entity{p}, where relevance is defined in terms of the extent to which the path is semantically related to \entity{p}. As an example, for the predicate \entity{director}, paths including predicates like \entity{director}, \entity{starring} \entity \textit{producer} are intuitively more relevant than paths including \entity{birthDate} or \entity{college}. To quantify the relevance  between a predicate and a schema-level pattern, the Path Extractor relies on a predicate relatedness measure, which given a pair of predicates $(p_i,p_j)$ computes their relatedness as:
\begin{equation}
Rel(p_i,p_j)=Cosine(Emb(p_i), Emb(p_j))
\label{eq:pred-rel}
\end{equation}

\noindent 
where $Emb(\cdot)$ is an embedding function (e.g., RotatE~\cite{sun2019rotate}) and \textit{Cosine} is the cosine operation between the vector embeddings of $p_i$ and $p_j$. Finally, the relatedness between a path and a predicate \entity{p} is computed as the average relatedness between \entity{p} and all predicates in the path.
%%%%%%%%%%%%%%%%%%%%%%%%%%%%%%
%%
\begin{algorithm}[!h]
	\scriptsize
	$\allEvPatterns=\emptyset$; /* priority queue based on 
	path relatedness  */\\
	$R$= \textit{getTopKPredicates}($\targetP$, $k$) \\
	$\domains$= \textit{getDomains}$(R)$ /* set of domains of predicates in R */ \\
	$\ranges$= \textit{getRanges}$(R)$ /* set of ranges of predicates in R */ \\
	\tcp{Parallel Execution:}
	\For{each predicate $p_i\in R$}
	{Let $\Delta_1=\emptyset$\\
		\For{$(t_r,p_i,t_s) \in  G_S$}{ 
			\If{$t_r\in \domains$ }{ 
				%$\Delta_f$=$buildEvidencePatterns(t_f)$\\
				$\Delta_1$=$\Delta_1 \cup \{p_i(t_r,t_s)\}$ \\
				\If{$t_s\in \ranges$}{	
					%	\If					{($checkTypes(p_i,\targetP)=\texttt{true})$}
					{$\allEvPatterns=\allEvPatterns\cup \{p_i(t_r,t_s),$\textit{getPathRelatedness}$(\targetP,p_i) \}$}}
			}
			\tcp{Same for $t_s\in \ranges$}
			%	\If{$t_s\in \ranges$}{
			%	$\Delta_1$=$\Delta_1 \cup \{(p_i(t_s,t_r)\}$\\
			%	\If{$t_r\in \domains$}{	
			%	{$\allEvPatterns=\allEvPatterns\cup\{p_i(t_s,t_r),$\textit{getRuleRelatedness}$(\targetP,p_i) \}$}}
			%}
		}
		\For{$j=2,...,d$}{
			Let $\Delta_{j}=\emptyset$;\\
			\For{each $\pi\in \Delta_{j-1}$}{
				%
				%	\For{$t_i$$\in$\textit{{expandBody}}$(\pi, G_S, R)$}{
				\For{$(t_i,p,t_j)$$\in$\textit{{expand}}$(\pi, G_S, R)$}{
					$\Delta_j=\Delta_j \cup \{\pi, (t_i, p,t_j)\}$ //add schema triple\\ 
				}
				%	\For{$(t_j,p,t_i)\in G_S$}{
				%	$\Delta_j=\Delta_j \cup \{\pi \wedge p(t_j,t_i)\}$ //add atom % 
				%
				%}
			}
			%}
			\For{each $\pi_j\in \Delta_j$}{
				%$\allEvPatterns=\allEvPatterns \cup\{\Pi_j, q_{rel}(\Pi_j) \}$
				\If 
				{\textit{checkRange}($\pi_j$,$\ranges$)}
				{$\allEvPatterns$=$\allEvPatterns\cup\{\pi_j,$  
					\textit{getPathRelatedness}$(\targetP,\pi_j) 
					\}$}
			}
		}
	}
	\Return $\allEvPatterns$
	\caption{\textit{extractSchemaLevelPatterns({\normalfont \entity{p}}$, k, d,G_S, M_R$)}}
	\label{alg:schema-level-paths}
\end{algorithm}
The relatedness-driven algorithm to extract schema-level patterns is sketched in Algorithm~\ref{alg:schema-level-paths}. At line 2 the algorithm extracts the top-$k$ predicates (controlled by the hyper-parameter $k$) related to \entity{p}. Then, for each such predicates it obtains the set of domains and ranges. This can be done in constant time by hashing the set of triples $T_t$ in the TBox related to the definition of domains and ranges of properties. The algorithm proceeds with a BFS traversal of the TBox graph conditioned on the top-$k$ predicates. The algorithm starts with  length-1 schema-level patterns (lines 8-11). At length $j<d$, the algorithm expands the last node found at $j-1$ (line 15) by getting both incoming and outgoing edges and nodes from the TBox graph definition. At line 17, the algorithm verifies that for each length $j< d$ the start and end node of the path are compatible with the domain and range of \entity{p}. Finally, the algorithm computes the path relatedness score (line 11 and line 19), which serves the purpose of inserting the schema-level patterns in a priority queue.

\noindent
\underline{Observations.}  We want to point out that schema-level patterns contain {both} entity types and predicates. As an example, for the predicate \entity{director}, the algorithm starts finding paths between \entity{Person} and \entity{Movie} or \entity{Director} and \entity{Movie}. As compared to meta-path based approaches (e.g.,~\cite{shi2016fact}) that use schema-level information to drive the extraction of information from the \kgraph\ data, our approach brings a twofold advantage. First, it can automatically extract and rank schema-level patterns based on their relevance with the input fact predicate \entity{p},while metapaths need to be manually defined. Second, it includes both entity types and predicates while metapaths only include entity types. It is important to note that including predicates allows to differentiate among multiple paths including the same entity types. Finally, observe that schema-level patterns are computed once per predicate and are used to compute the final data-level paths depending on the specific pair of entities (\entity{s}, \entity{o}) that replace the patterns' endpoints.

\subsubsection{Data-level paths.}The  Path Extractor has available a set of schema-level patterns $\mathcal{P}_\entity{p}$, for each predicate \entity{p}, found in the previous step. Hence, given an input fact \fact{s}{p}{o}, the goal is to find data-level paths from the ABox for each schema-level pattern $\pi_i \in \mathcal{P}_\entity{p}$. This is done by relying on an algorithm based on a variant of Depth-First-Search (DFS), which starts from \entity{s} and at each traversal step of the graph ensures the compliance with $\pi_i$ in terms of predicate traversed and entity types toward reaching the entity \entity{o}. Consider the fact \fact{Dune}{director}{D. Lynch}, the schema-level-path {\scriptsize$\pi=\rightPathElement{Work}{starring}{Actor}$$\leftPathElement{}{starring}{Work}$$\rightPathElement{}{director}{Person}$} and the \kgraph\ in Fig. \ref{fig:example-graph} (a). The algorithm starts from the node \entity{Dune}  and traverses the edge \entity{starring} (as per $\pi$) reaching the nodes \entity{J. Nance}, \entity{K. Mclaughlin}, and \entity{E. McGill}. From each of these nodes, it traverses edges labeled as \entity{starring} in reverse direction (again as per $\pi$) and reaches the nodes \entity{Eraserhead}, \entity{Twin Peaks} and  \entity{Twin Peaks Fire Walks with Me}. Finally, according to the last step of $\pi$, the algorithm traverses edges labeled as \entity{director} thus closing the paths between the subject \entity{Dune} and the object \entity{ D. Lynch} of the input fact. Note that when considering the pattern {\scriptsize $\pi=\rightPathElement{Film}{cinematography}{Person}$$\leftPathElement{}{director}{Film}$$\rightPathElement{}{editing}{Person}$}, it is not possible to find any path between \entity{Dune} and \entity{D. Lynch} complying with $\pi$ in the ABox. If no path complying with any schema-level pattern can be found, \system\ performs an unconstrained DFS.
\subsection{Path Embedder}
\label{sec:path-embedder}
To be processed by the learning model at the core of \system, paths found by the Path Extractor are given a numerical representation. This is done by vectorizing each fact (triple) in a path, which can be done in different ways. One way is to consider techniques like TransE~\cite{bordes2013translating} or DistMult~\cite{yang2015embedding} to first learn entity and predicate embeddings via a generic function \emb{$\cdot$}, which given an entity or a predicate, returns its corresponding vector embedding. Hence, to compute the embedding of a fact \entity{t}=\fact{s}{p}{o}, one can perform some operation $op$ (e.g., concatenation) on its constituents vectors, that is, \emb{t}=$op$(\emb{s},\emb{p},\emb{o}). Note that we do not consider one-hot encodings since these techniques do not take into account the structure of the \kgraph. Another way to learn fact embeddings is to rely on approaches like {triple2vec}~\cite{fionda2019triple2vec}, which instead of learning embeddings for entities and predicates separately directly learns fact embeddings. We will report on the performance of \system\ when considering both approaches in Section~\ref{sec:large-scal-eval}.
For the time being, given a fact $t$=$\factDef$, we define its embedding as $t_E$=$\embF{t}$.
Building upon the embedding of facts, a path $\pi$=\{$t^1,t^2,\ldots t^l$\} of length $l$ including $l$ facts is encoded as a sequence $\pi_E$=[$t_E^1,t_E^2,\ldots,t_E^l$].}

We observe that previous attempts to encode paths (e.g., \cite{agrawal2019learning}) focused on single label graphs only intending to tackle the link prediction problem in terms of predicting whether a link, \textit{no matter the specific predicate}, exists between a pair of nodes. Our goal in this paper is to consider \kgraphs\ including different entity and predicate types. Doing so it is possible to explicitly incorporate the semantics of predicates into path representations, which
paves the way to applications like fact-checking, where the goal is to establish whether a specific relation expressed, via a predicate, exists between a pair of nodes.
%%%%
\subsection{Path Aggregator}
\label{sec:path-aggregator}
Paths converted into their vector form by the Path Embedder are then passed to the Path Aggregator. The Aggregator implements a variety of aggregation strategies (detailed Section~\ref{sec:aggregators}). At this stage, we can see the aggregator as another learning module, which takes the paths from the Path Embedder and provides an overall vector representation for them. As the Path Extractor groups paths according to their different lengths, the Path Aggregator processes each length-specific set of paths separately. Finally, the path representations for each length are concatenated together to give the final length-specific path representation $\mathcal{P}_V$ (see Fig. \ref{fig:learning-model}).
%%%%
\subsection{Fact Checker}
\label{sec:fact-checker}
%%%%
The last step of the \system\ framework consists in providing the final truthfulness score about the input fact. This is done by the Fact Checker, which takes as input the output of the Path Aggregator (i.e., the vector representation $\mathcal{P}_V$) and feeds it into a classifier. We treat the fact-checking problem as a binary classification problem, where a true fact and a false fact are assigned 1 and 0 as target values, respectively. The final goal is to optimize the negative log-likelihood objective function, which defined  as follows:
\begin{equation}
\mathcal{L}=- \sum_{f^+  \in \mathcal{F}^+}   log \ \hat{y}_{f^+} + \sum_{f^-  \in \mathcal{F}^-}log(1- \ \hat{y}_{f^-})
\end{equation}

where $\mathcal{F}^+$=\{$f^+\mid y_{f^+}=1$\} and $\mathcal{F}^-$=\{$f^+\mid y_{f^-}=0$\} are the true and false facts, respectively. 
\section{Aggregators}
\label{sec:aggregators}
In the previous sections, we have outlined the \system\ architecture. In particular, we have discussed how \system\ from a set of paths interlinking \entity{s} and \entity{o} (obtained by the Path Extractor) obtains an overall vector representation via the Path Aggregator. The idea of aggregating graph information has been previously used (e.g., GraphSage~\cite{hamilton2017inductive}, LEAP \cite{agrawal2019learning})  mainly for node classification and link prediction, where the goal is to predict whether a link exists between a pair of nodes no matter the label. However, these pieces of related work have  only considered nodes in a graph disregarding edges, edge labels, and the \kgraph\ schema.

Our goal is to aggregate path information involving both nodes and labeled edges that are crucial for fact-checking using \kgraphs, where the goal is to establish whether a specific semantic relation  between a pairs of nodes holds. In particular, we are concerned with aggregating sequences of paths of the form [\fact{s$_1$}{p$_1$}{o$_1$}, \fact{o$_1$}{p$_2$}{o$_3$}, $\ldots$ \fact{o$_n$}{p$_n$}{o$_{n+1}$}] instead of paths intended as sequences of nodes [\entity{s$_1$},\entity{o$_1$},\entity{o$_3$},$\ldots$ \entity{o$_n$},\entity{o$_{n+1}$}]. Another crucial difference between \system\ and related approaches (e.g., LEAP \cite{agrawal2019learning}) is the fact that the Path Extractor module only considers paths relevant to the specific domain of the fact to be checked (expressed via the fact predicate), while these related work extract all paths for link prediction. The  intuition is that to (dis)prove a fact, a subset of domain-related paths can provide the necessary pieces of semantic evidence.
In general, we can see an instance of Aggregator as a neural network that takes as input the paths (along with the batch size, fact embeddings, number of paths, path length) and gives a final vector representation. Inspired by previous work \cite{agrawal2019learning}, we considered three main aggregation strategies.
%%%%%%%
%%%%%
\subsection{Average Pool}
\label{sec:avg-pool-aggr}
This kind of aggregator is the simplest we consider. It combines the different representations of paths by concatenating the vector representations of the facts in a path. Then on the set of paths obtained, the  aggregator performs a 1D average pooling operation. The final combined path representation is a single vector obtained  by averaging the paths between \entity{s} and \entity{o}. The whole operation can be summarized as follows:
%%%
\begin{equation}
\mathcal{P}_V^l={AvgPool([\oplus(\pi_i^l), \forall \pi_i^l \in \mathcal{P}^l])}
\end{equation}

where $AvgPool$ is the one-dimensional average pooling operation, and $\oplus(\cdot)$ is the vector concatenation operation. This representation relies on the embeddings of the facts in each path.
\subsection{Max Pool}
\label{sec:sense-max-aggregator}
This kind of aggregator shares with the AvgPool the fact representation obtained, even in this case, by concatenating the vectors of both the nodes and the predicate  in a fact. What changes is the final vector of the path; instead of being the average, it is now computed by using a dense neural network layer. The resulting activations are then passed through a max-pooling operation which helps to derive a single vector representation for the paths of length $l$. The whole operation can be summarized as follows:
\begin{equation}
\mathcal{P}_V^l={MaxPool([ \sigma(W_l\cdot\oplus(\pi_i^l)+b_l), \forall \pi_i^l \in \mathcal{P}^l])}
\end{equation}

where $MaxPool$ is the one-dimension max pool operation (which selects bitwise the maximum value from multiple vectors to derive a single final vector.), $W_l$ are the weights to be learned, $b_l$ the bias, and $\sigma$ the activation function.
\subsection{LSTM Max Pool}
\label{sec:recurrent-aggregator}
We now outline the most sophisticated aggregator we considered. The idea is to treat a (vectorized) path as a sequence an employ an LSTM network to cater for sequential dependencies between facts in a path. With this reasoning, each fact in a path represents a point of a sequence. At each step $l-1$,  the  LSTM layer outputs a hidden state vector $h_{l-1}$, consuming subsequence of embedded facts $[f_1,..., f_{l-1}]$. In other words, $x_{l-1}$=$f_{l-1}$. The input $x_{l-1}$ and the hidden state $h_{l-1}$ are used to learn the hidden state of the next path step $l$. 
As our final goal is to leverage the representations of all paths, after processing all of them via the LSTM, the aggregator employs another LSTM followed by a max pool operation to produce the combined path representation $\mathcal{P}_V$.
%%%
%%%%%%%%
\section{Evaluation}
\label{sec:impl-eval}
%%%%
We designed \system\ with a modular architecture, which makes it easy to consider different strategies for embedding facts and different aggregation mechanisms.
We tested the ability of our approach to check facts considering both existing and not existing facts in a given \kgraph\ on synthetic (Section \ref{sec:large-scal-eval}) and existing benchmarks (Section \ref{sec:existing-benchmarks}).  Details about the experimental setting, the datasets, and the implementation are available in the Appendix.
\subsection{Evaluation Methodology}
To evaluate the performance of \system\ and competitors, we use the Area Under the Receiver Operating Characteristic curve (AUC). We identify AUC as the primary quality indicator because is a generic measure (independent of the threshold values) and has been extensively used in previous work (e.g., \cite{shiralkar2017finding,fionda2018fact}). All experiments have been carried out on a machine with a 4 core 2.7 GHz CPU and 16 GB RAM.

%%%
\begin{figure*}[t]
	\centering
\includegraphics[width=.7\textwidth, angle =-90 ]{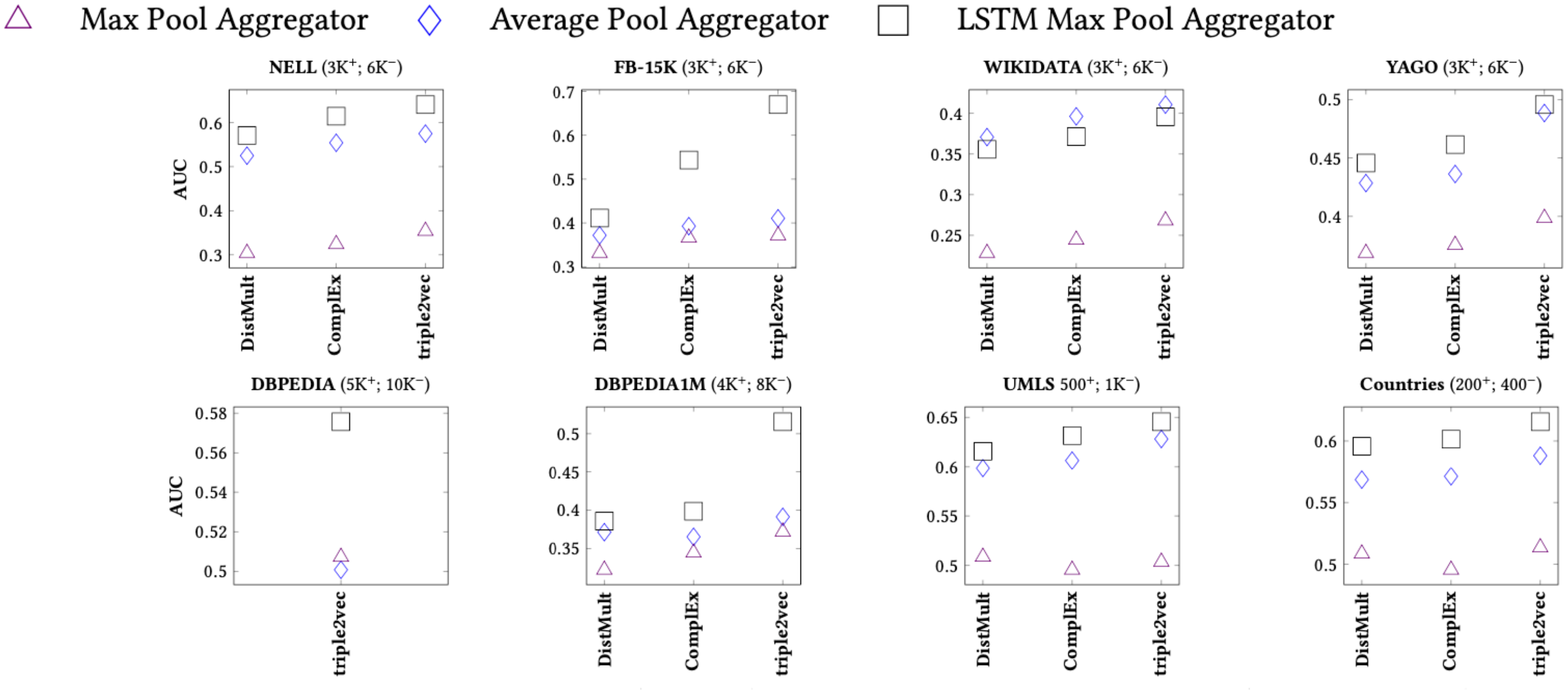}
	%%%
	\vspace{-2.3cm}
	\caption{{For each dataset, $+$ (resp., $-$) denotes the number of positive (resp., negative) examples.}}
	\label{fig:embeding-method}
\end{figure*}
\begin{figure*}[t]
\centering
\includegraphics[width=.7\textwidth, angle =-90 ]{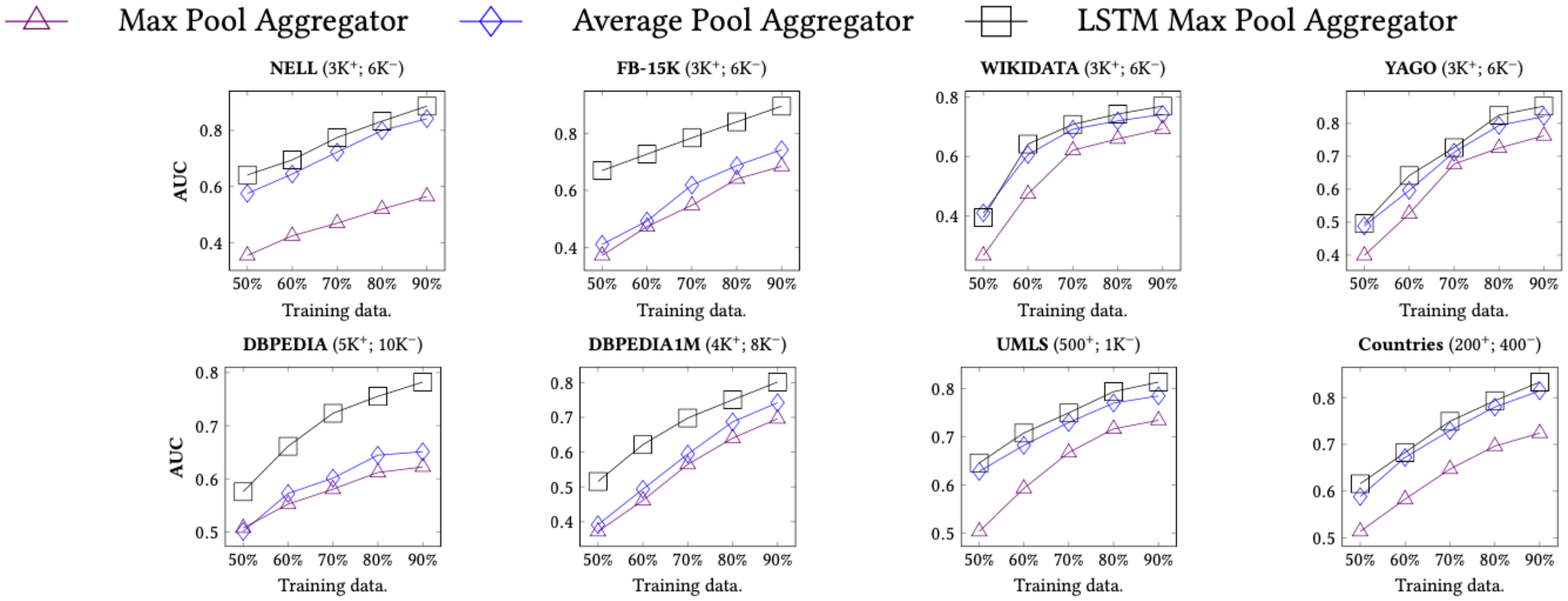}
	\vspace{-2.5cm}
	\caption{AUC values for increasing values of training data.}
	\label{fig:split-variation}
\end{figure*}
\subsection{Embedding and Relatedness Computation} 
 We considered three different approaches to pre-compute fact embeddings. \textbf{DistMult} \cite{yang2015embedding} and \textbf{ComplEx} \cite{trouillon2016complex} give embeddings for entities and predicates separately; in this case, the embedding of a fact is obtained as the \textit{concatenation} of the embeddings of its constituent parts. The third approach, \textbf{triple2vec} \cite{fionda2019triple2vec} directly computes fact embeddings by leveraging the notion of line graph of a \kgraph. In particular, the triples of a \kgraph\ become the nodes of the line graph. The approach then uses random walks and the word2vec approach to find the embeddings of each node of the line graph, which gives the embeddings of the triples of the original graph.

Predicate embeddings were used to obtain a predicate relatedness matrix, where the relatedness of each pair of predicates is computed as per equation~(\ref{eq:pred-rel}) for all datasets but DBpedia. In this case, we obtained the predicate relatedness matrix from KStream\footnote{\url{https://github.com/shiralkarprashant/knowledgestream}}. This was necessary since neither DistMult nor ComplEx could run on this dataset on our machine.
 %%%
\subsection{Large scale evaluation}
\label{sec:large-scal-eval}
We conducted a large scale fact-checking experiment on all \kgraphs\ by generating synthetic benchmarks. In this experiment, \textit{the goal was to test the impact of model parameters like the approach to compute fact embeddings, train and test size and, most important, aggregation strategy}. Facts for training and testing were extracted to cover all the different predicates proportionally (i.e., the more popular a predicate the larger the number of triples for fact-checking). It is important to observe that positive facts used in these experiments were removed from the \kgraph\ during the evaluation. The generation of train and test instances is discussed in the Appendix.
%%%%

\subsubsection{Impact of fact embedding} We evaluated the impact of the fact embedding approach used by the Path Embedder when considering the three aggregation strategies. The number of positive and negative examples considered are shown close to the dataset name in Fig. \ref{fig:embeding-method}. Note that the number of false facts considered is double than that of true facts. This is to take into account the fact that there are many more possible false statements than there  are  true  statements in the synthetic generated datasets. Moreover, we considered 50\% of these examples (both positive and negative) for training and the remaining for testing. We observe that in almost all datasets \textbf{triple2vec} provides better performance. This is especially true as the size of the \kgraph\ increases. For instance, in the smallest dataset Countries, we observe that \textbf{triple2vec} coupled with the LSTM Aggregator gives the best performance, even if \textbf{DistMul} and \textbf{CompleEx} performed almost equally good when considering this aggregator. For DBpedia, we only report results for \textbf{triple2vec} as \textbf{DistMul} and \textbf{ComplEx} raised an out-of-memory error.  In general, we observe that with the  \textit{AvgPool} and  \textit{MaxPool} aggregators the impact of the embedding strategy is less significant. 

We hypothesize that the better performance of \textbf{triple2vec} is because it computes embeddings \textit{for every fact} while for the other two approaches fact embeddings often share entities and predicate representations. Moreover, the LSTM can better capture dependencies between triples in the fact when considering the specific embedding of each  triple. Neither this could be achieved with \textbf{DistMult} nor with \textbf{ComplEx} that both compute unique predicate/entity embeddings that will be used in all facts in which a predicate/entity participates into. Although the \textit{LSTMMaxPool} performed better than the other two, we note that Wikidata is an exception where the Avg aggregator provided slightly better results. Perhaps this is due to the way the dataset was selected, which only considers facts from the top-100 most popular predicates.
\begin{table*}[]
	\renewcommand\arraystretch{0.05}
	\centering
	\small
	\begin{tabular}{@{}cccccccccccc@{}}
		\toprule
		\textbf{Approach}                   & \textbf{birthPlace}   & 
		\textbf{deathPlace}    & \textbf{almaMater}     & 
		\textbf{nationality}  & 
		\textbf{profession} & \multirow{7}{*}{} & \textbf{author}   
		& 
		\textbf{team}        & \textbf{director}     & 
		\textbf{keyPerson}   & 
		\textbf{spouse}     \\
		\textbf{}                   & (273/1092)   & 
		(126/504)     & (1546/6184)      & 
		(50/200)   & 
		(110/440)  & \multirow{7}{*}{} & (93/558)  
		& 
		(41/164)        & (78/4680)   & 
		(201/1208)   & 
		(16/256)     \\
		\cmidrule(r){1-6}\cmidrule(l){8-12}  \system-LSTMAggr &	\multicolumn{1}{c}{.93} 
		& 
		\multicolumn{1}{c} {.91} & \multicolumn{1}{c}{.81} & 
		\multicolumn{1}{c}{.89} & 
		\multicolumn{1}{c}{.99} &                   & 
		\multicolumn{1}{c}{.93} & 
		\multicolumn{1}{c} {.92} & \multicolumn{1}{c}{.99} & 
		\multicolumn{1}{c}{.84} & 
		\multicolumn{1}{c} {.98} \\ 
		\cmidrule(r){1-6} \cmidrule(l){8-12} 
		\cmidrule(r){1-6}\cmidrule(l){8-12}  \system-MaxAggr &	\multicolumn{1}{c}{.90} 
		& 
		\multicolumn{1}{c} {.87} & \multicolumn{1}{c}{.80} & 
		\multicolumn{1}{c}{.86} & 
		\multicolumn{1}{c}{.97} &                   & 
		\multicolumn{1}{c}{.90} & 
		\multicolumn{1}{c} {.91} & \multicolumn{1}{c}{.97} & 
		\multicolumn{1}{c}{.79} & 
		\multicolumn{1}{c} {.94} \\ 
		\cmidrule(r){1-6} \cmidrule(l){8-12} 
		\cmidrule(r){1-6}\cmidrule(l){8-12}  \system-AvgAggr &	\multicolumn{1}{c}{.91} 
		& 
		\multicolumn{1}{c} {.86} & \multicolumn{1}{c}{.81} & 
		\multicolumn{1}{c}{.87} & 
		\multicolumn{1}{c}{.99} &                   & 
		\multicolumn{1}{c}{.92} & 
		\multicolumn{1}{c} {.93} & \multicolumn{1}{c}{.99} & 
		\multicolumn{1}{c}{.81} & 
		\multicolumn{1}{c} {.98} \\ 
		\cmidrule(r){1-6} \cmidrule(l){8-12} 
		\multicolumn{1}{l}{CHEEP}& 
		\multicolumn{1}{c}{.91} 
		& 
		\multicolumn{1}{c} {.87} & \multicolumn{1}{c}{.77} & 
		\multicolumn{1}{c}{.85} & 
		\multicolumn{1}{c}{.98} &                   & 
		\multicolumn{1}{c}{.91} & 
		\multicolumn{1}{c} {.91} & \multicolumn{1}{c}{.99} & 
		\multicolumn{1}{c}{.82} & 
		\multicolumn{1}{c} {.96} \\ \cmidrule(r){1-6} \cmidrule(l){8-12} 
		\multicolumn{1}{l} {KStream}  &
		\multicolumn{1}{c}{.82} 
		& 	\multicolumn{1}{c}{.84} & \multicolumn{1}{c}{.75} & 
		\multicolumn{1}{c}{.93} 
		& 
		\multicolumn{1}{c}{.93} &                 & \multicolumn{1}{c}{.92} & 
		\multicolumn{1}{c}{.99} & \multicolumn{1}{c}{{.83}} & 
		\multicolumn{1}{c}{.80} & 
		\multicolumn{1}{c}{{.86}} \\ \cmidrule(r){1-6} \cmidrule(l){8-12} 
		\multicolumn{1}{l}{KLinker}  & 
		\multicolumn{1}{c}{.91} 
		& 
		\multicolumn{1}{c}{.87} & \multicolumn{1}{c}{.78} & 
		\multicolumn{1}{c}{.86} 
		& 
		\multicolumn{1}{c}{.93} &                   & \multicolumn{1}{c}{.96} & 
		\multicolumn{1}{c}{.92} & \multicolumn{1}{c}{{.88}} & 
		\multicolumn{1}{c}{.83} & 
		\multicolumn{1}{c}{{.91}} \\ \cmidrule(r){1-6} 
		\cmidrule(l){8-12} 
		\multicolumn{1}{l}{PredPath} & \multicolumn{1}{c}{.86} & 
		\multicolumn{1}{c}{.76} & \multicolumn{1}{c}{.83} & 
		\multicolumn{1}{c}{.95} & 
		\multicolumn{1}{c}{.92} &                   & \multicolumn{1}{c}{.99} & 
		\multicolumn{1}{c}{.92} & \multicolumn{1}{c}{{.84}} & 
		\multicolumn{1}{c}{.88} & 
		\multicolumn{1}{c}{{.87}} \\  \cmidrule(r){1-6} 
		\cmidrule(l){8-12} 
		\multicolumn{1}{l}{PRA} & \multicolumn{1}{c}{.74} & 
		\multicolumn{1}{c}{.75} & \multicolumn{1}{c}{.63} & 
		\multicolumn{1}{c}{.83} & 
		\multicolumn{1}{c}{.50} &                   & \multicolumn{1}{c}{.96} & 
		\multicolumn{1}{c}{.91} & \multicolumn{1}{c}{{.99}} & 
		\multicolumn{1}{c}{.87} & 
		\multicolumn{1}{c}{{.88}} \\  \cmidrule(r){1-6} 
		\cmidrule(l){8-12} 
		\multicolumn{1}{l}{{LEAP}} & \multicolumn{1}{c}{.81} & 
		\multicolumn{1}{c}{.74} & \multicolumn{1}{c}{.80} & 
		\multicolumn{1}{c}{.91} & 
		\multicolumn{1}{c}{.88} &                   & \multicolumn{1}{c}{.99} & 
		\multicolumn{1}{c}{.89} & \multicolumn{1}{c}{{.81}} & 
		\multicolumn{1}{c}{.82} & 
		\multicolumn{1}{c}{{.85}} \\  \cmidrule(r){1-6} 
		\cmidrule(l){8-12} 
		\multicolumn{1}{l}{TransE}& \multicolumn{1}{c}{.54} & 
		\multicolumn{1}{c}{.56} & \multicolumn{1}{c}{.66} & 
		\multicolumn{1}{c}{.77} & 
		\multicolumn{1}{c}{.82} &                   & \multicolumn{1}{c}{.80} & 
		\multicolumn{1}{c}{.56} & \multicolumn{1}{c}{{.82}} & 
		\multicolumn{1}{c}{.83} & 
		\multicolumn{1}{c}{{.79}} \\ 
		\bottomrule
	\end{tabular}
	\caption{{Performance (average AUC) on both real-world (left) 
		and synthetic (right) datasets (average of 4 runs).} }
	%The number of true 
	%facts over all facts is reported below the predicate name.
	\label{tab:results-AUC}
	
	\vspace{-.8cm}
\end{table*}

\subsubsection{Impact of Train/Test Split}
In this experiment, we report results only when considering \textbf{triple2vec} as a fact embedding approach. We also tested the other two fact embedding approaches but they gave an inferior performance on all datasets. We set the hyper-parameter $k$=10,  which controls the number of predicates for the generation of schema-level patterns (see Section \ref{sec:schema-level-paths}).

Fig. \ref{fig:split-variation} reports results on all datasets. We observe that the performance improves as more training data are used. Nevertheless, already with only 50\% of training data, the AUC score is consistently above 0.5 in all datasets when considering the  \textit{LSTMMaxPool} aggregator. Observe that this kind of aggregator gave the best performance for all different train/test splits. Nevertheless, in smaller datasets like UMLS and Countries, the  \textit{AvgPool} aggregator also gave good results.
We also experimented with other values of $k$ (results are not reported for lack of space) and observed that: (i) for broad-domain \kgraphs\ (e.g., DBpedia, YAGO, NELL)) a value higher than 10 does not bring any improvement as  actually it worsen the performance. This is due to the fact that paths collected include predicates that do not contribute to a precise contextualization of the fact; (ii) for more specific \kgraphs\ like UMLS and Countries, larger values of $k$ offer better performance; even if when considering more than 30\% of all available predicates, the performance starts to decrease. Overall, we observed that a value of $k$  between 7 and 10 offers the best trade-off between the time required to find schema and data-level paths and AUC score. Results on more specific datasets like UMLS and Countries are better in general. 
This can be explained by the fact that the paths used as evidence to (dis)prove a fact offer a clearer support for the learning model; in other words, while in more general \kgraphs\ there can be paths that can lead the model astray in terms of fact contextualization, in more specific datasets this is less probable. It is interesting to observe that thanks to the hyper-parameter $k$, \system\ can control how much it can be led astray from the domain expressed by \entity{p}. This offers greater flexibility than previous work where it is not possible to contextualize targeted facts in terms of paths generated.
%%%
%%
%%
\subsection{Comparison with related work}
\label{sec:existing-benchmarks}
Building upon the analysis conducted in Section~\ref{sec:large-scal-eval}, we now report on the comparison of \system\ with related work on existing benchmarks.
 We considered: \textit{(i)}  \textbf{CHEEP}~\cite{fionda2018fact}, an approach, which leverages paths to come up with a truthfulness score for an input fact; \textit{(ii)} \textbf{PredPath} ~\cite{shi2016discriminative}, which exploits frequent anchored predicate paths between pair of entities in the \kgraph; \textit{(iii)} Path Ranking Algorithm (\textbf{PRA}) \cite{gardner2014incorporating}, which extracts (positive and negative) training set of triples via a two-sided {unconstrained} random walk starting from the fact endpoints to retrieve paths between them; \textit{(iv)} \textbf{KStream}~\cite{shiralkar2017finding}, which reduces the fact-checking problem to the problem of maximizing the flow between the subject and the object of the fact; \textit{(v) } \textbf{Klinker} ~\cite{ciampaglia2015computational}, which relies  on  a  single  short,  specific  path  to differentiate  between a true and a false fact; \textit{(vi)} \textbf{LEAP}~\cite{agrawal2019learning}, which tackles the problem of link prediction on unlabeled graphs. We included LEAP as we took inspiration from it for the path aggregation strategies, although \system\ tackles the more challenging problem of fact-checking. For LEAP, paths were generated ignoring the edge labels and we report the best results obtained with its aggregation strategies; \textit{(vii)} we could not run experiments with DistMult and ComplEx because of memory issues. Nevertheless, we report the results for \textbf{TransE}~\cite{bordes2013translating} obtained by Shiralkar et al. \cite{shiralkar2017finding}.
 We did not consider approaches that can check facts via logical rules learned from the \kgraph\ (e.g.,~\cite{leblay2017declarative}) since it is not completely clear how to obtain high-quality rules.
 %%
%%%
 \subsubsection{Benchmarks} We compared the various systems on two benchmarks. The first defined in Shiralkar et al.~\cite{shiralkar2017finding} and available online\footnote{\url{https://github.com/shiralkarprashant/knowledgestream/}}. It includes 5 real-world datasets derived from Google Relation Extraction Corpora and WSDM Cup Triple Scoring challenge and 5 synthetic datasets mix a-priori known true and false facts. As the benchmarks are defined on DBpedia entities, we considered this \kgraph\ as a source of background knowledge in this experiment. The number of true/false facts for each benchmark is reported below the predicate name in Table \ref{tab:results-AUC}. 
 The second benchmark\footnote{\url{https://github.com/huynhvp/BUCKLE-Fact_checking/}} released by Huynh and Papotti \cite{huynhbenchmark2019} takes into account popularity, transparency, homogeneity, and functionality properties of the facts to cover a broader variety of scenarios than previous benchmarks.
 
 \subsubsection{Evaluation Results} We observe that on the first benchmark (Table~\ref{tab:results-AUC}), \textbf{\system}\ performs quite well for all predicates considered. In particular, it brings some improvement wrt \textbf{CHEEP}, the second-best performing system, when considering the  \textit{LSTMMaxPool} aggregator. We note that approaches like \textbf{PredPath}, which only consider one path perform worse; perhaps a single path is not able to capture all needed semantic evidence. As expected, the worst-performing system is \textbf{LEAP}, which, however, has not been designed to work on labeled graphs as it aims to solve the link prediction problem in unlabeled graphs. We included it since it also features path aggregation strategies that were a source of inspiration in designing \system. However, the semantics of predicates and paths seems to play a crucial role in fact-checking. 
 
 We observe that \textbf{TransE}, which also tackles the link prediction problem, but on knowledge graphs, performs better than \textbf{LEAP}; although worse than the other systems. This could be because it does not consider paths. Results are more interesting in the second benchmark (Table \ref{tab:eval-2}), which has carefully been designed to test the behavior of fact-checking systems on (non)popular (NP) and random entities (R). On this benchmark, we ran experiments for \textbf{\system}, \textbf{LEAP}, and \textbf{CHEEP} while for \textbf{TransE}, \textbf{KLinker}, and \textbf{PredPath} we report results from \cite{huynhbenchmark2019}. Here we observe that \textbf{\system}\ performs particularly well on non-popular entities with both  \textit{LSTMMaxPool} and  \textit{Avg} aggregators. This may be explained by the fact that even when the number of paths is smaller than between popular entities, the Path Aggregator can correctly capture the necessary evidence, which passed to the other modules of the \textbf{\system}\ framework (after embedding) captures the truthfulness of facts eventually. 

\textbf{\system} leverages deep-learning techniques for the embedding of paths providing a strategy that can capture dependencies between the facts in a path and aggregate them. Moreover, we remark the importance of considering the semantics of paths for fact-checking but, most importantly, the need to correctly relate the semantics of such paths with the fact to be checked in order to only consider the most relevant ones. Indeed, even if \textbf{LEAP} uses node embedding and path aggregation strategies, it is the worst performing system. We noted that the system fails to especially recognize false facts since even if the existence of a link is correctly predicted, this is not enough as for fact-checking it is necessary to establish the existence of a specific link.
%%%
\vspace{.4cm}
%%%
\begin{table}[]
	\small
		\renewcommand\arraystretch{.5}
	\begin{tabular}{cclll}
		\textbf{Approach}             & \multicolumn{1}{c|}{\textbf{Predicate}}    & \multicolumn{1}{c|}{\textbf{P}} & \multicolumn{1}{c|}{\textbf{NP}} & \textbf{R} \\ \hline
		\multirow{4}{*}{FEA-LSTMAggr} & \multicolumn{1}{c|}{\textbf{nearestCity}}  & \multicolumn{1}{c|}{.87}              & \multicolumn{1}{c|}{.67}                  &  .78               \\ \cline{2-5} 
		& \multicolumn{1}{c|}{\textbf{foundedBy}}    & \multicolumn{1}{l|}{.82}                 & \multicolumn{1}{l|}{.67}                     &   .86              \\ \cline{2-5} 
		& \multicolumn{1}{c|}{\textbf{manufacturer}} & \multicolumn{1}{l|}{.91}                 & \multicolumn{1}{l|}{.88}                     &      .94           \\ \cline{2-5} 
		& \multicolumn{1}{c|}{\textbf{employer}}     & \multicolumn{1}{l|}{.70}                 & \multicolumn{1}{l|}{.52}                     &  .71               \\ \hline      \hline     
		%%%%%%
		\multirow{4}{*}{FEA-MaxAggr} & \multicolumn{1}{c|}{\textbf{nearestCity}}  & \multicolumn{1}{c|}{.79}              & \multicolumn{1}{c|}{.61}                  &  .72               \\ \cline{2-5} 
		& \multicolumn{1}{c|}{\textbf{foundedBy}}    & \multicolumn{1}{l|}{.77}                 & \multicolumn{1}{l|}{.63}                     & .79                \\ \cline{2-5} 
		& \multicolumn{1}{c|}{\textbf{manufacturer}} & \multicolumn{1}{l|}{.88}                 & \multicolumn{1}{l|}{.79}                     &    .86             \\ \cline{2-5} 
		& \multicolumn{1}{c|}{\textbf{employer}}     & \multicolumn{1}{l|}{.66}                 & \multicolumn{1}{l|}{.47}                     & .62                \\ \hline      \hline   
		%%%%
		\multirow{4}{*}{FEA-AvgAggr} & \multicolumn{1}{c|}{\textbf{nearestCity}}  & \multicolumn{1}{c|}{.85}              & \multicolumn{1}{c|}{.64}                  &     .75            \\ \cline{2-5} 
		& \multicolumn{1}{c|}{\textbf{foundedBy}}    & \multicolumn{1}{l|}{.79}                 & \multicolumn{1}{l|}{.63}                     &       .80          \\ \cline{2-5} 
		& \multicolumn{1}{c|}{\textbf{manufacturer}} & \multicolumn{1}{l|}{.89}                 & \multicolumn{1}{l|}{.88}                     &     .91            \\ \cline{2-5} 
		& \multicolumn{1}{c|}{\textbf{employer}}     & \multicolumn{1}{l|}{.68}                 & \multicolumn{1}{l|}{.50}                     &  .67               \\ \hline      \hline   
		%%%%
		\multirow{4}{*}{CHEEP} & \multicolumn{1}{c|}{\textbf{nearestCity}}  & \multicolumn{1}{c|}{.86}              & \multicolumn{1}{c|}{.61}                  &     .72            \\ \cline{2-5} 
		& \multicolumn{1}{c|}{\textbf{foundedBy}}    & \multicolumn{1}{l|}{.81}                 & \multicolumn{1}{l|}{.62}                     &   .79              \\ \cline{2-5} 
		& \multicolumn{1}{c|}{\textbf{manufacturer}} & \multicolumn{1}{l|}{.78}                 & \multicolumn{1}{l|}{.86}                     &      .81           \\ \cline{2-5} 
		& \multicolumn{1}{c|}{\textbf{employer}}     & \multicolumn{1}{l|}{.67}                 & \multicolumn{1}{l|}{.43}                     &   .64              \\ \hline      \hline  
				\multirow{4}{*}{PredPath} & \multicolumn{1}{c|}{\textbf{nearestCity}}  & \multicolumn{1}{c|}{.84}              & \multicolumn{1}{c|}{.58}                  &  .69               \\ \cline{2-5} 
		& \multicolumn{1}{c|}{\textbf{foundedBy}}    & \multicolumn{1}{l|}{.80}                 & \multicolumn{1}{l|}{.63}                     &    .81             \\ \cline{2-5} 
		& \multicolumn{1}{c|}{\textbf{manufacturer}} & \multicolumn{1}{l|}{.55}                 & \multicolumn{1}{l|}{.51}                     &    .53             \\ \cline{2-5} 
		& \multicolumn{1}{c|}{\textbf{employer}}     & \multicolumn{1}{l|}{.58}                 & \multicolumn{1}{l|}{.38}                     &  .50               \\ \hline      \hline  
		% 
%		\multirow{4}{*}{KStream} & \multicolumn{1}{c|}{\textbf{nearestCity}}  & \multicolumn{1}{c|}{.87}              & \multicolumn{1}{c|}{.66}                  &      .76           \\ \cline{2-5} 
%		& \multicolumn{1}{c|}{\textbf{foundedBy}}    & \multicolumn{1}{l|}{.82}                 & \multicolumn{1}{l|}{.67}                     &  .80               \\ \cline{2-5} 
%		& \multicolumn{1}{c|}{\textbf{manufacturer}} & \multicolumn{1}{l|}{.90}                 & \multicolumn{1}{l|}{.85}                     & .92                \\ \cline{2-5} 
%		& \multicolumn{1}{c|}{\textbf{employer}}     & \multicolumn{1}{l|}{.69}                 & \multicolumn{1}{l|}{.43}                     & .66                \\ \hline      \hline   
		%     
			\multirow{4}{*}{KLinker} & \multicolumn{1}{c|}{\textbf{nearestCity}}  & \multicolumn{1}{c|}{.87}              & \multicolumn{1}{c|}{.66}                  &      .76           \\ \cline{2-5} 
		& \multicolumn{1}{c|}{\textbf{foundedBy}}    & \multicolumn{1}{l|}{.82}                 & \multicolumn{1}{l|}{.67}                     &  .80               \\ \cline{2-5} 
		& \multicolumn{1}{c|}{\textbf{manufacturer}} & \multicolumn{1}{l|}{.90}                 & \multicolumn{1}{l|}{.85}                     & .92                \\ \cline{2-5} 
		& \multicolumn{1}{c|}{\textbf{employer}}     & \multicolumn{1}{l|}{.69}                 & \multicolumn{1}{l|}{.43}                     & .66                \\ \hline      \hline   
			\multirow{4}{*}{LEAP} & \multicolumn{1}{c|}{\textbf{nearestCity}}  & \multicolumn{1}{c|}{.41}              & \multicolumn{1}{c|}{.40}                  & .41                \\ \cline{2-5} 
		& \multicolumn{1}{c|}{\textbf{foundedBy}}    & \multicolumn{1}{l|}{.69}                 & \multicolumn{1}{l|}{.58}                     & .71                \\ \cline{2-5} 
		& \multicolumn{1}{c|}{\textbf{manufacturer}} & \multicolumn{1}{l|}{.68}                 & \multicolumn{1}{l|}{.57}                     & .64                \\ \cline{2-5} 
		& \multicolumn{1}{c|}{\textbf{employer}}     & \multicolumn{1}{l|}{.58}                 & \multicolumn{1}{l|}{.42}                     &      .43           \\ \hline      \hline   
			\multirow{4}{*}{TransE} & \multicolumn{1}{c|}{\textbf{nearestCity}}  & \multicolumn{1}{c|}{.49}              & \multicolumn{1}{c|}{.40}                  &     .43            \\ \cline{2-5} 
		& \multicolumn{1}{c|}{\textbf{foundedBy}}    & \multicolumn{1}{l|}{.75}                 & \multicolumn{1}{l|}{.60}                     &  .75               \\ \cline{2-5} 
		& \multicolumn{1}{c|}{\textbf{manufacturer}} & \multicolumn{1}{l|}{.72}                 & \multicolumn{1}{l|}{.47}                     &   .70             \\ \cline{2-5} 
		& \multicolumn{1}{c|}{\textbf{employer}}     & \multicolumn{1}{l|}{.62}                 & \multicolumn{1}{l|}{.46}                     &  .48               \\ \hline      \hline   
	\end{tabular}
\caption{{AUC results on the benchmarks in \cite{huynhbenchmark2019} for popular (\textbf{P}), non popular (\textbf{NP}), and random (\textbf{R}) entity pairs. Train and test pairs has  been provided by the authors of \cite{huynhbenchmark2019}.}}
\label{tab:eval-2}
\end{table}
%%%
%%%
\section{Related Work}
\label{sec:rel-work}
%%%%
Several approaches have been proposed to leverage structured information from knowledge graphs to (dis)prove facts of the form \fact{s}{p}{o}. The common approach is to consider the set of paths interlinking \entity{s} and \entity{o} as  a form of evidence. 

CHEEP \cite{fionda2018fact} leverages the schema to generate evidence patterns that are then used to generate data-level paths whose "support" is then used to provide a final truthfulness score. Although we share the usage of the schema, our approach can automatically learn and weight the importance of the different paths, including the sequence of facts in a path, thanks to an end-to-end modular framework. KStream \cite{shiralkar2017finding} casts the fact-checking problem into that of computing the maximum flow between \entity{s} and \entity{o}. Although we share the usage of predicate relatedness, KStream does not use it to prune the path search space as we do. Moreover,  KStream does not consider the \kgraph\ schema and thus the computation of the flow can include a large portion of the \kgraph. Finally, our approach can automatically learn and aggregate path representations thanks to the usage of embedding and deep learning techniques. {PredPath}~\cite{shi2016discriminative} focuses on finding a single path that can shed light on the truthfulness of a fact. As resulted from our evaluation, a single path may be unable to correctly single out the necessary body of semantic evidence.   LEAP \cite{agrawal2019learning} uses path aggregation strategies and node embeddings to predict the existence of a link between a pair of input nodes. Although we took inspiration from LEAP to define the path aggregation strategies, our approach differs in several respects. First, LEAP focuses on shorter paths while we focus on paths that are semantically relevant to the input fact. Second, LEAP works on unlabeled graphs while we considered labeled multi-graphs. As such, LEAP is unable to predict whether a specific link exists between a pair of nodes, which is the goal of fact-checking systems. Indeed, checking the fact \fact{Dune}{director}{D. Lynch} is different from checking the existence of a link between \entity{Dune} and \entity{D. Lynch}. 
Our approach to generate schema-level patterns shares some commonalities with rule learning systems (e.g.,~\cite{galarraga2015fast}) although our goal is to only consider patterns that are relevant to the input fact. Moreover, our work feds data-level paths to a deep-learning pipeline for embedding and aggregation. Our work also differs from logic-based approaches (e.g.,~\cite{leblay2017declarative}) aiming at representing/querying facts to capture incompleteness/uncertainty. 

There are a variety of link prediction systems, including those based on embeddings (e.g.,~\cite{shi2017proje,trouillon2016complex,yang2015embedding}), that can be applied to the task of fact-checking. Our approach differs in several respects. First, our model goes beyond embedding-based approaches that only leverage entity embeddings. \system's learning model is based on triple (fact) and path embeddings, and their aggregation, including a mechanism (based on LSTMs) to capture sequential dependencies between facts in a path, and a weighted pooling mechanism that weights their relative importance toward the final verdict. Moreover, \system\ can provide evidence for a fact thanks to the set of paths input to the learning model that can be visualized. In a way this allows to combine the benefits of pure path-based approaches (e.g., \cite{shiralkar2017finding,fionda2018fact}) with those of deep-learning models. Other systems leverage text (e.g., \cite{,ercan2019retrieving,syed2018factcheck}) and are out of the scope of this analysis, which is focused on how to embed and carefully aggregate paths from \kgraphs\ for fact checking.
\section{Conclusions and Future Work}
\label{sec:conclusions}
%%%%
Knowledge Graphs can provide useful support, in terms of structured background knowledge, for fact-checking. Our proposal aims at bridging the gap between two existing strands of approaches, that is, path-based approaches and embedding based approaches each with its relative merits. \system\ achieves this goal thanks to the interplay between a schema-based algorithm, which carefully focuses on the subset of paths most relevant to an input fact, and a path embedding/aggregation mechanism able to distillate this body of semantic evidence. 
Our experiments showed that the fact embedding mechanism plays an important role; the one-size-fits-all entity/predicate embedding approach (e.g., TransE~\cite{bordes2013translating}, DistMult~\cite{yang2015embedding}) is sub-optimal as compared to approaches that directly embed each fact (i.e., triple2vec~\cite{fionda2019triple2vec}) in a path. It also emerged that capturing the dependencies between facts (via the \textit{LSTMMaxPool} aggregator) can improve the performance. Overall, fact-checking cannot be solved by looking at the fact alone; it is necessary to understand its semantics, contextualize it by considering related (chains of) facts, and distillate semantic evidence from paths. Investigating other aggregation strategies, the usage of adversarial learning techniques~\cite{goodfellow2014generative}, and  temporal information~\cite{ercan2019retrieving} is in our research agenda.
%%%%%%%%
\bibliographystyle{ACM-Reference-Format}
\bibliography{fact-checking-paths} 

\newpage
\onecolumn
\section*{Appendix}
\subsection{Hyperparameters}
 We considered the hyperparameter $l$=\{1,2, 3, 4\} (path lengths) and considered the top-50 schema level paths (stored in the priority queue as per Algorithm~\ref{alg:schema-level-paths} line 1). For each such schema-level pattern, we used up to 150 data-level paths for each length randomly selected. We used the Adam \cite{kingma2014adam} optimizer with learning rate set to 0.001 and trained the model for 100 epochs with early stopping enabled.
%%%
%%%%
\subsection{Knowledge graphs and datasets}
We performed experiments using several  \kgraphs\ as a source of background knowledge for fact-checking. Details about the datasets are available in Table~\ref{tab:datasets}. 
The datasets are taken from a variety of real-world \kgraphs, some of which model broad knowledge (e.g., DBpedia, Yago) while others are more domain-specific (e.g., UMLS, Countries). We considered a subset of Wikidata sampling from the 100 predicates that have the most number of facts. We considered the version of DBpedia in~\cite{shiralkar2017finding} but discarded typing information. We also considered a subset of it (DBpedia1M). For Countries and UMLS that do not have a schema, \system\ extracted data-level paths by performing a constrained DFS, which only retrieves paths involving the top-$k$ most related input predicates to a targeted predicate.
%%%%%%%
\begin{table}[!h]
	\renewcommand\arraystretch{1.2}
	\begin{tabular}{|c|c|c|c|}
		\hline
		\textbf{Dataset} & \textbf{\#Entities} & \textbf{\#Relations} & \textbf{\#Facts} \\ \hline
		{Countries}\tablefootnote{\url{https://github.com/shehzaadzd/MINERVA}}        & $\sim$250         & 2                  & $\sim$1K       \\ \hline
		{UMLS} $^7$            & $\sim$130         & 49                 & $\sim$5K       \\ \hline
		%	\texttt{Kingship} (\texttt{KGS})         & $\sim$100         & 26                 & $\sim$17K      \\ \hline
		{NELL-995}  $^7$         & $\sim$75K         & 200                & $\sim$155K     \\ \hline
		{FB15K-237}$^7$        & $\sim$14.5K       & 237                & $\sim$272K     \\ \hline
		{Yago}\tablefootnote{\url{https://github.com/yago-naga/}}         & $\sim$3.5K        & 37                 & $\sim$355K     \\ \hline
		{Wikidata}\tablefootnote{\url{{https://www.wikidata.org}}}       & $\sim$100K        & 100                & $\sim$698K     \\ \hline
		DBpedia\tablefootnote{\url{http://carl.cs.indiana.edu/data/}}       & $\sim$4M          & 661                & $\sim$8M      \\ \hline
		{DBpedia1M}      & $\sim$276K          & 2284                & 1M      \\ \hline
	\end{tabular}
	\caption{{Knowledge graphs considered.}}
	\label{tab:datasets}
	\vspace{-.9cm}
\end{table}
%%%%%%%%
%%%%%%%%%%%%%%%%%%%%
\subsection{Training and test instances generation}
\label{sec:ex-generation}
Existing fact-checking benchmarks (e.g., \cite{shiralkar2017finding,huynhbenchmark2019}) have been defined on DBpedia. In order to evaluate \system\ on a larger variety of \kgraphs, we generated synthetic benchmarks including positive and negative (\entity{s}, \entity{o}) pairs for each predicate \entity{p} in a \kgraph. Note that we treat \kgraphs\ as trusted sources of knowledge (i.e., facts are assumed to be correct) but incomplete because of the open-world assumption, which states that a fact not in the \kgraph\ can either be false or missing.
Let $\mathcal{T}^+_\entity{p}$=\{(\entity{s}, \entity{o}): \fact{s}{p}{o} $\in G$\} be the set of pairs linked by \entity{p} existing in a knowledge graph $G$ and let $\mathcal{T}^-_\entity{p}$=\{(\entity{s}, \entity{o'}): \fact{s}{p}{o'} $\not\in G$\} the set of pairs that are not linked by \entity{p} in $G$. Given a predicate \entity{p}, to collect \textit{positive train instances}, a subset of pairs $\mathcal{T}_{tr}^+ \subset \mathcal{T}^+$ is used. In particular, for each schema-level pattern obtained for \entity{p}, each pair (\entity{s}, \entity{o})$\in \mathcal{T}_{tr}^+$ replaces the schema-level pattern's endpoints and allows to find a set of data-level paths. The remaining set of pairs $\mathcal{T}_{ts}^+ =\mathcal{T}^+ \setminus  \mathcal{T}_{tr}^+$ represent the \textit{positive test pairs}.
To collect negative pairs, we adopt the Local Closed World Assumption (LCWA), which states that if a \kgraph\ contains one or more object (resp. subject) values for a given subject (resp. object) and predicate, then it contains all possible occurrences for the two entities involved for that predicate~\cite{galarraga2015fast,huynhbenchmark2019}. Moreover, to ensure high accuracy for the generated negative pairs, the types of the entities are required to be the same in a new negative pair. For example, if the fact \entity{D. Lynch} \entity{director} \entity{Dune}, it implies that the \kgraph\ contains all information about the director of \entity{Dune}. Therefore, for any other director (e.g., \entity{S. Kubrick}) in the \kgraph, its combination with \entity{Dune} is a false fact (i.e., \fact{S. Kubrick}{director}{Dune} is a false fact). Note that to generate more precise negative examples the type of the entity that we replace needs to be the same (i.e., a director). We also divide the negative train ($\mathcal{T}_{tr}^- \subset \mathcal{T}^-$) and test ( $\mathcal{T}_{ts}^- =\mathcal{T}^- \setminus  \mathcal{T}_{tr}^-$) pairs.
%%%
%%
\subsection{Reproducibility}
%%%
%%%
\system\footnote{{The system and the datasets are available upon request}} has been implemented in Python using the GraphTool library\footnote{\url{https://graph-tool.skewed.de/}} to deal with graph data and pathfinding, pykg2vec\footnote{\url{https://pypi.org/project/pykg2vec/}} to compute entity and predicate embeddings, and KERAS/Tensorflow\footnote{\url{https://keras.io}} for the implementation of the aggregators, the training and testing of the whole system. 
\end{document}